\documentclass{article}

    \IfFileExists{neurips_2024.sty}{%
      \usepackage{neurips_2024}
    }{%
      \usepackage[margin=1.2in]{geometry}
    }

    \usepackage[utf8]{inputenc}
    \usepackage[T1]{fontenc}
    \usepackage[protrusion=true,expansion=false]{microtype}
    \usepackage{hyperref}
    \usepackage{url}
    \usepackage{booktabs}
    \usepackage{amsmath, amsfonts, amssymb}
    \usepackage{graphicx}
    \usepackage[section]{placeins}
    \usepackage{xcolor}
    \usepackage{tikz}
    \usetikzlibrary{positioning,arrows.meta,plotmarks,calc}
    \usepackage{pgfplots}
    \usepgfplotslibrary{groupplots}
    \pgfplotsset{compat=1.18}

    \newcommand{\TCAFIGDIR}{../figures}
    \IfFileExists{../figures/triad_diagram.tikz.tex}{}{%
      \IfFileExists{figures/triad_diagram.tikz.tex}{\renewcommand{\TCAFIGDIR}{figures}}{%
        \IfFileExists{triadic_cognitive_architecture/figures/triad_diagram.tikz.tex}{\renewcommand{\TCAFIGDIR}{triadic_cognitive_architecture/figures}}{}%
      }%
    }%

    \graphicspath{{figures/}{../figures/}{../../figures/}{../../../figures/}{triadic_cognitive_architecture/paper/figures/}{triadic_cognitive_architecture/figures/}{../triadic_cognitive_architecture/figures/}{../../triadic_cognitive_architecture/figures/}{../../../triadic_cognitive_architecture/figures/}}

    \newcommand{\TCAIncludeGraphics}[2][]{%
      \IfFileExists{\TCAFIGDIR/#2}{%
        \includegraphics[#1]{\TCAFIGDIR/#2}%
      }{%
        \IfFileExists{figures/#2}{\includegraphics[#1]{figures/#2}}{%
        \IfFileExists{../figures/#2}{\includegraphics[#1]{../figures/#2}}{%
          \IfFileExists{../../figures/#2}{\includegraphics[#1]{../../figures/#2}}{%
            \IfFileExists{../../../figures/#2}{\includegraphics[#1]{../../../figures/#2}}{%
              \IfFileExists{triadic_cognitive_architecture/paper/figures/#2}{\includegraphics[#1]{triadic_cognitive_architecture/paper/figures/#2}}{%
                \IfFileExists{triadic_cognitive_architecture/figures/#2}{\includegraphics[#1]{triadic_cognitive_architecture/figures/#2}}{%
                  \IfFileExists{../triadic_cognitive_architecture/figures/#2}{\includegraphics[#1]{../triadic_cognitive_architecture/figures/#2}}{%
                    \IfFileExists{../../triadic_cognitive_architecture/figures/#2}{\includegraphics[#1]{../../triadic_cognitive_architecture/figures/#2}}{%
                      \IfFileExists{../../../triadic_cognitive_architecture/figures/#2}{\includegraphics[#1]{../../../triadic_cognitive_architecture/figures/#2}}{%
                        \typeout{TCA: Missing figure: #2}%
                        \fbox{\texttt{Missing figure: \detokenize{#2}}}%
                      }%
                    }%
                  }%
                }%
              }%
            }%
          }%
        }}%
      }%
    }

    \newcommand{\TCAInputTikZ}[1]{%
      \IfFileExists{\TCAFIGDIR/#1}{%
        \input{\TCAFIGDIR/#1}%
      }{%
        \IfFileExists{figures/#1}{\input{figures/#1}}{%
        \IfFileExists{../figures/#1}{\input{../figures/#1}}{%
          \IfFileExists{../../figures/#1}{\input{../../figures/#1}}{%
            \IfFileExists{../../../figures/#1}{\input{../../../figures/#1}}{%
              \IfFileExists{triadic_cognitive_architecture/paper/figures/#1}{\input{triadic_cognitive_architecture/paper/figures/#1}}{%
                \IfFileExists{triadic_cognitive_architecture/figures/#1}{\input{triadic_cognitive_architecture/figures/#1}}{%
                  \IfFileExists{../triadic_cognitive_architecture/figures/#1}{\input{../triadic_cognitive_architecture/figures/#1}}{%
                    \IfFileExists{../../triadic_cognitive_architecture/figures/#1}{\input{../../triadic_cognitive_architecture/figures/#1}}{%
                      \IfFileExists{../../../triadic_cognitive_architecture/figures/#1}{\input{../../../triadic_cognitive_architecture/figures/#1}}{%
                        \typeout{TCA: Missing TikZ: #1}%
                        \fbox{\texttt{Missing TikZ: \detokenize{#1}}}%
                      }%
                    }%
                  }%
                }%
              }%
            }%
          }%
        }}%
      }%
    }

    \title{Cognitive Friction: A Decision-Theoretic Framework for Bounded Deliberation in Tool-Using Agents}

    \author{Davide Di Gioia \\ \texttt{ucesigi@ucl.ac.uk}}
    \date{}

\begin{document}

    \maketitle

    \begin{abstract}
    Autonomous tool-using agents operating in networked environments must decide both \emph{which} information source to query and \emph{when} to stop querying and act. Without principled bounds on information-acquisition costs, unconstrained agents exhibit systematic failure modes: excessive tool use under congestion, prolonged deliberation under time decay, and brittle behavior under ambiguous evidence. We propose the Triadic Cognitive Architecture (TCA), a decision-theoretic framework that formalizes these failure modes through \textit{Cognitive Friction}. By synthesizing nonlinear filtering theory, congestion-dependent cost dynamics, and HJB optimal stopping, we model deliberation as a stochastic control problem over a joint belief--congestion state space, where information acquisition is explicitly priced by tool-dependent signal quality and live network load. Rather than relying on heuristic stop-tokens or fixed query budgets, TCA yields an HJB-inspired stopping boundary and a computable rollout-based approximation of belief-dependent value-of-information with a net-utility halting condition. We validate TCA on two controlled simulation environments, the Emergency Medical Diagnostic Grid (EMDG) and the Network Security Triage Grid (NSTG), designed to isolate stopping quality, action selection under congestion, and temporal urgency under reproducible conditions. TCA reduces time-to-action while improving resource outcomes without degrading accuracy: over greedy baselines, TCA gains 36 viability points in EMDG and 33 integrity points in NSTG, and purposive stopping baselines confirm that stopping rules alone recover at most 4 viability points. Finally, we provide an illustrative instantiation around a black-box LLM on a memorisation-free corpus, showing that the same stopping principle executes using empirically computable uncertainty and VOI proxies.
    \end{abstract}

    \section{Introduction}

    Autonomous agents that interact with networked environments through tool queries (whether in medical triage, network security, or open-ended task execution) face a fundamental decision problem: \emph{which} information source to query next, and \emph{when} to stop querying and act. Iterative query-then-decide loops such as ReAct \cite{yao2023react}, Tree-of-Thoughts \cite{yao2023tot}, and AutoGPT \cite{autogpt2023} have demonstrated that sequential querying substantially improves decision quality. However, these systems operate under the implicit assumption of a \textit{frictionless cognitive environment}: gathering information is presumed to be instantaneous, structurally free, and uniformly reliable.

    When deployed outside of static text-generation benchmarks and into dynamic, physical, or highly networked environments, this assumption mathematically can lead to systematic failure modes. Without intrinsic bounds on spatial topology, temporal pacing, or epistemic resolution, unconstrained agents exhibit three critical failure modes:
    \begin{enumerate}
      \item \textbf{Congestion Saturation (Space):} Indiscriminate API querying cascades into network congestion and rising per-query load costs, as agents fail to price the structural cost of routing information through a multi-agent swarm or distributed database.
      \item \textbf{Infinite Deliberation (Time):} Lacking a continuous-time pacing mechanism, agents get trapped in recursive thought loops, failing to recognize that the utility of an optimal decision decays exponentially as time elapses.
      \item \textbf{Epistemic Collapse (Truth):} When faced with unresolvable, contradictory evidence, agents lack a rigorous mathematical framework for doubt. Instead of exposing the ambiguity, they average conflicting claims and confidently hallucinate a hybrid synthesis.
    \end{enumerate}

    In this work, we argue that robust autonomous control requires a formal, decision-theoretic optimization layer that prices information acquisition explicitly. We introduce the \textbf{Triadic Cognitive Architecture (TCA)}, which models agent deliberation as a bounded stochastic control problem over a joint belief-routing state space, in which information acquisition is explicitly priced by spatio-temporal and epistemic costs.

    Building on related ideas in congestion-aware routing, interval-aware decision-making, and entropic claim resolution \cite{digioia2026cascade,digioia2026learning,digioia2026entropic}, we formally couple the belief state of an agent to the physical friction of its environment. We present a continuous-time formulation for generality (nonlinear filtering and HJB-inspired stopping); in our EMDG implementation we instantiate beliefs on a finite hypothesis set and approximate value-of-information by Monte Carlo rollouts (Section~\ref{sec:emdg_discrete}).

    \paragraph{Novelty and contributions.}
    We contribute: (i) an idealized continuous-time stochastic control formulation in which bounded deliberation is characterized by an HJB-inspired stopping boundary; (ii) a computable discrete instantiation, with formal approximation error bounds (Propositions~1--2) bounding the discrete-to-continuous gap via It\^{o}'s formula and Hoeffding's inequality, that estimates belief-dependent VOI via rollouts and halts by net utility; (iii) an empirical evaluation on EMDG and a second distinct environment (NSTG) showing that cost-aware selection and principled stopping jointly drive consistent $30{+}$-point improvements over greedy baselines; and (iv) a continuation-value sweep confirming that $\eta{=}0$ is optimal on EMDG trajectories under high temporal urgency.

    The remainder of this paper is structured as follows. Section 2 reviews related work, anchoring TCA within classical bounded rationality and modern agentic AI. Section 3 formalizes the physics of Cognitive Friction and derives the triadic stochastic control objective. Section 4 empirically validates the framework via the Emergency Medical Diagnostic Grid (EMDG). Finally, Section 5 discusses the implications of TCA as a foundational framework for safe, scalable autonomy.

    \section{Related Work}

    The Triadic Cognitive Architecture bridges classical theories of bounded computation with modern tool-using autonomous agent design. We briefly contextualize our contributions within these intersecting domains.

    \subsection{Tool-Using Autonomous Agents}
    Sequential tool-use and query-then-decide loops have been widely adopted in modern agentic architectures. Frameworks such as ReAct \cite{yao2023react}, Tree-of-Thoughts \cite{yao2023tot}, and Reflexion \cite{shinn2023reflexion} have demonstrated that interleaving reasoning with environmental observations significantly improves task success rates. However, these frameworks rely on arbitrary stopping constraints (e.g., token limits or maximum step counts) \cite{wang2023survey}, treating each query as free regardless of latency or congestion. Evaluation benchmarks such as AgentBench \cite{liu2023agentbench} effectively quantify agent failure modes in interactive settings, but implicitly treat deliberation computation as a free resource. TCA argues that safe autonomy demands that computation itself be optimized as a structural cost.

    \subsection{Bounded Rationality and Meta-Reasoning}
    The necessity of pricing deliberation traces back to Simon's foundational concept of bounded rationality \cite{simon1955behavioral}, which posits that agents must make decisions under the strict limits of available information, cognitive capacity, and time. This was formally extended into meta-reasoning and bounded optimality by Russell and Wefald \cite{russell1991right}, who argued that an agent should compute only until the cost of computation outweighs the expected increase in decision utility. In deep learning, Adaptive Computation Time (ACT) \cite{graves2016adaptive} introduced halting mechanisms for recurrent networks based on internal confidence. TCA extends these classical concepts to tool-using autonomous agents, formalizing sequential tool queries as a continuous-time physical process subject to network congestion and temporal decay.
    Crucially, classical meta-reasoning and VOI-based planners typically treat information-acquisition costs as static scalars independent of agent state.  TCA instead makes costs \emph{dynamically state-dependent}: spatial friction grows with live congestion $C_t$ and temporal friction accumulates with delay $t$, both changing as deliberation proceeds.  This coupling of state-dependent cost dynamics with belief-dependent VOI within a single HJB stopping framework is the key structural departure from prior bounded rationality work.

    \subsection{Active Inference and Stochastic Control}
    Our framework shares philosophical DNA with Karl Friston’s Free Energy Principle and Active Inference \cite{friston2010free}, as well as Yann LeCun's Joint Embedding Predictive Architecture (JEPA). Both theories propose that intelligent agents interact with their environment to minimize surprise (or expected free energy) via a predictive world model. While foundational, exact Bayesian active inference remains largely computationally intractable in practice. TCA offers a principled and computable alternative by providing a closed-loop stochastic control envelope. By synthesizing non-linear filtering \cite{kushner1964differential, bain2009fundamentals} with congestion-aware routing and HJB optimal stopping \cite{shiryaev2007optimal, oksendal2003stochastic}, TCA bridges stochastic control theory with practical agent systems engineering.

    \section{Formalizing Cognitive Friction: The Coupled Stochastic Control Objective}
    \paragraph{Problem setup.}
    We consider an agent that acquires information via costly tool queries and must decide both which query to perform next and when to stop deliberating and execute a terminal decision. The agent maintains a belief state over hypotheses and a congestion state tracking accumulated network load.

    \paragraph{Intuition (for ML practitioners).}
    At inference time, a tool-using agent faces two coupled questions: \emph{which tool should I query next?} and \emph{when should I stop querying and act?}
    Most agent loops answer both with heuristics (fixed budgets, step limits, or ad hoc confidence thresholds).
    TCA instead prices deliberation explicitly: each prospective query has (i) a belief-dependent expected benefit (value-of-information (VOI): expected entropy reduction) and (ii) an environment-dependent cost (latency and congestion).
    The controller chooses the action maximizing net utility and halts when even the best available query has non-positive net value.
    The continuous-time HJB formulation provides the principled stopping boundary; EMDG instantiates a computable discrete approximation via rollout-based VOI and a myopic stopping rule.

    We define the active cognitive state at time $t$ as $s_t=(p_t,C_t)$, where $p_t\in\Delta^{|\Theta|-1}$ is the agent's belief over a finite hypothesis set $\Theta$, and $C_t\in\mathbb{R}_{\ge 0}$ is the scalar network congestion state (accumulated load). Selecting action $u_t\in\mathcal{A}$ increments congestion by the action's load cost:
    \begin{equation}
    \dot{C}_t = \Omega(u_t),
    \end{equation}
    where $\Omega:\mathcal{A}\to\mathbb{R}_{\ge 0}$ is the load-increment function.

    \subsection{The observation model and belief dynamics}
    The agent acquires evidence through noisy tool observations. The chosen action $u_t\in\mathcal{A}$ selects which channel to query; each tool produces a different signal profile under each hypothesis. The observation process $Y_t$ follows an It\^o diffusion:
    \begin{equation}
    dY_t = h(\theta,u_t)\,dt + \sigma\,dW_t,
    \end{equation}
    where $h(\theta,u_t)\in\mathbb{R}^d$ is the expected signal of tool $u_t$ under hypothesis $\theta$, $W_t\in\mathbb{R}^d$ is standard Brownian motion, and $\sigma>0$ is a fixed positive scalar so the noise covariance is $\sigma^2 I_d$. Observation quality depends on which tool is selected but not on congestion; congestion enters exclusively through the cost of acquiring observations (Section~\ref{sec:value_function}).
    The belief state $p_t$ evolves via the Wonham (finite-state Kushner--Stratonovich) filter, driven by the innovation process $d\nu_t = dY_t - \bar{h}(p_t,u_t)\,dt$, where $\bar{h}(p_t,u_t)=\int_\Theta h(\theta,u_t)\,p_t(\theta)\,d\theta$ is the belief-weighted mean signal of the selected tool.

    \subsection{The triadic value function}
    \label{sec:value_function}
    The agent seeks to maximize expected information gain while minimizing spatio-temporal cognitive friction. We define two explicit, state-dependent cost terms:
    \begin{align}
    \ell_S(C_t, u_t) &:= \lambda_S\bigl(C_t + \Omega(u_t)\bigr), \nonumber\\
    \ell_T(t) &:= \beta\cdot t. \nonumber
    \end{align}
    Here $\lambda_S>0$ is a fixed spatial cost weight. The spatial cost $\ell_S$ is \emph{state-dependent}: it grows with the live congestion $C_t$ and the load increment $\Omega(u_t)$ of the chosen action.
    The temporal cost $\ell_T$ reflects the opportunity cost of delay. We use $\beta t$ as a linearly increasing surrogate for time urgency; this aligns directly with the discrete implementation, where the temporal friction term is $\beta(t+\tau(a))$ (a per-action linear penalty). The exponential decay in the resource metric ($e^{-\beta\tau(a)/100}$ per query in EMDG) is a consequence of accumulated linear penalties and does not require an exponential integral formulation.
    The value function is defined as the supremum over stopping times $T\ge t$:
    \begin{equation}
    V(p,C,t) = \sup_{T\ge t}\, \mathbb{E}\!\left[\,\int_t^{T}\!\left(\frac{1}{2\sigma^2}\,\mathbb{E}_{p_s}\!\left[\bigl\|h(\theta,u_s)-\bar{h}(p_s,u_s)\bigr\|^2\right] - \lambda_S\bigl(C_s+\Omega(u_s)\bigr) - \beta\,s\right)ds\;\middle|\; p_t=p,\;C_t=C\,\right],
    \end{equation}
    where $\bar{h}(p_s,u_s)$ is the belief-weighted mean signal (Section~3.1) and $\sigma$ is the scalar observation noise (Eq.~2); the term $\frac{1}{2\sigma^2}\mathbb{E}_{p_s}[\|h(\theta,u_s)-\bar{h}(p_s,u_s)\|^2]$ equals the expected instantaneous entropy-reduction rate of the Wonham filter. Different tools yield different $h(\theta,u)$ profiles; the $\sup_{u\in\mathcal{A}}$ in the HJB below selects the tool that maximises net reward, trading off information gain against cost.
    The state $(p_t,C_t,t)$ is the minimal sufficient statistic for future net reward: $p_t$ determines expected information gain and $C_t$ determines the cost trajectory of future queries.

    \subsection{The HJB variational inequality and optimal stopping}
    Let $\mathcal{L}^u$ denote the (controlled) infinitesimal generator of the Markov state $(p_t,C_t)$ under control $u\in\mathcal{A}$, incorporating:
    (i) the Wonham (finite-state Kushner--Stratonovich) filtering dynamics in $p_t$ (see \cite{kushner1964differential,bain2009fundamentals}); and
    (ii) the deterministic congestion dynamics $\dot{C}_t=\Omega(u)$.
    By dynamic programming, $V(p,C,t)$ satisfies the Hamilton--Jacobi--Bellman (HJB) variational inequality:
    \begin{equation}
    0 = \max\left\{\,\partial_t V(p,C,t) + \sup_{u\in\mathcal{A}}\!\left(\mathcal{L}^u V(p,C,t) + \frac{1}{2\sigma^2}\,\mathbb{E}_{p}\!\left[\bigl\|h(\theta,u)-\bar{h}(p,u)\bigr\|^2\right] - \lambda_S(C+\Omega(u)) - \beta\,t\right),\; -V(p,C,t)\,\right\}.
    \end{equation}
    In the discrete EMDG instantiation, the running temporal penalty is approximated by an action-dependent one-step cost proportional to latency $\tau(a)$ (Section~\ref{sec:emdg_discrete}).

    \subsection{Discrete-time instantiation used in EMDG (rollout VOI + myopic stopping)}
    \label{sec:emdg_discrete}
    While the preceding subsections present an idealized continuous-time formulation (filtering and HJB-inspired stopping),
    the EMDG experiments instantiate a discrete approximation aligned with tool-using agents.

    \paragraph{Notation.}
    We use \(\alpha\) for cost-to-utility scaling, \(\beta\) for temporal decay/latency penalty, and \(\lambda_S\) for the spatial cost weight; \(\Omega(a)\) and \(\tau(a)\) denote action load and latency.
    Value-of-information (VOI) is defined as belief-dependent expected entropy reduction, i.e., \(\widehat{\Delta H}(a\mid b_t)=\mathbb{E}[H(b_t)-H(b_{t+1})\mid a]\), estimated by rollouts in EMDG.
    In the continuous-time formulation $\beta(t)$ is time-varying; in EMDG we approximate it by a constant per-step rate $\beta$ applied to action latency $\tau(a)$.
    The logged quantity \texttt{info\_gain} denotes the realized one-step entropy reduction $H(b_t)-H(b_{t+1})$; total \texttt{info\_gain} telescopes to $H(b_0)-H(b_T)$.
    In the EMDG implementation, the continuous-time running reward $\frac{1}{2\sigma^2}\mathbb{E}_p[\|h(\theta,u)-\bar{h}(p,u)\|^2]$ is replaced by expected entropy reduction, estimated directly via rollout Bayesian updates: each rollout clones the current belief, samples a Dirichlet-likelihood observation, performs the Bayesian update, and records the entropy drop $H(b_t)-H(b_{t+1})$.

    \begin{table}[htbp]
    \centering
    \caption{Notation summary for the discrete EMDG instantiation.}
    \label{tab:notation}
    \small
    \begin{tabular}{ll}
    \toprule
    Symbol & Meaning \\
    \midrule
    $b_t \in \Delta^{|\Theta|-1}$ & Categorical belief over hypotheses at step $t$ \\
    $H(b_t)$ & Shannon entropy of belief $b_t$ \cite{shannon1948} \\
    $\widehat{\Delta H}(a \mid b_t)$ & Rollout-estimated VOI (expected entropy reduction) \\
    $\alpha$ & Cost-to-utility scaling (\texttt{cost\_scale}) \\
    $\beta$ & Per-step temporal decay/latency penalty (\texttt{beta}) \\
    $\lambda_S$ & Spatial cost weight (\texttt{lambda\_s}) \\
    $C_t$ & Current network congestion at step $t$ \\
    $\Omega(a)$ & Load increment of action $a$ (\texttt{load\_cost}) \\
    $\tau(a)$ & Latency of action $a$ (\texttt{time\_cost}) \\
    $U(a;\,b_t,t,C_t)$ & Net utility of action $a$ (stop when $\max_a U \le 0$) \\
    \texttt{info\_gain} & Realized one-step entropy reduction $H(b_t)-H(b_{t+1})$ \\
    \bottomrule
    \end{tabular}
    \end{table}

    We estimate the VOI term $\widehat{\Delta H}(a\mid b_t)$ via rollouts.
    The belief state is a categorical distribution $b_t\in\Delta^{|\Theta|-1}$ over a finite hypothesis set.
    Given an action $a$ (a tool query) and an observation sample, we update $b_t$ via Bayes' rule and compute categorical entropy
    $H(b)=-\sum_i b(i)\log\big(b(i)+\varepsilon\big)$.

    To estimate belief-dependent value-of-information, we use Monte Carlo rollouts on cloned states:
    \begin{equation}
    \widehat{\Delta H}(a \mid b_t)
    = \frac{1}{K}\sum_{k=1}^{K}\Big(H(b_t)-H(b_{t+1}^{(k)}(a))\Big).
    \label{eq:voi_rollout}
    \end{equation}
    This rollout estimator is a Monte Carlo approximation of the mutual information $I(\Theta;Y\mid a,b_t)$ between hypotheses and observations under the tool's observation model, a standard information-theoretic quantity that is well-defined for discrete hypothesis sets without requiring a Fisher information matrix.
    We then score actions by a net-utility objective that prices congestion and latency:
    \begin{equation}
    U(a; b_t, t, C_t)
    = \widehat{\Delta H}(a \mid b_t)
    -\alpha\Big(\lambda_S(C_t+\Omega(a)) + \beta\,(t+\tau(a))\Big),
    \label{eq:net_utility}
    \end{equation}
    and stop when the best continuation utility is non-positive:
    \begin{equation}
    \texttt{STOP}\;\Longleftrightarrow\;\max_{a\in\mathcal{A}}U(a; b_t, t, C_t)\le 0.
    \label{eq:stop_rule}
    \end{equation}
    Note that the triadic controller optimizes a proxy objective: expected entropy reduction minus spatio-temporal friction. The physical resource metrics reported in our experiments (patient viability and system integrity) are external environmental variables that decay with elapsed time. By explicitly pricing delay through the temporal term $\beta\,(t+\tau(a))$, the controller is designed to trade off information gain against preservation of these external metrics without requiring them to appear directly in the agent's internal belief state.
    This myopic stopping rule omits continuation value (equivalently, setting a one-step lookahead weight $\eta=0$ in a one-step lookahead formulation); we empirically validate the optimality of this choice in Section~\ref{sec:continuation}.
    Here, $C_t$ denotes the congestion state (Eq.~1), while $\eta$ is a separate scalar lookahead weight (set to zero in our experiments).

    \paragraph{On the discretization gap.}
    The idealized HJB variational inequality optimizes a continuation value over future information paths.
    Our EMDG controller uses a myopic rule ($\eta=0$), i.e., it compares immediate VOI against immediate spatio-temporal cost and halts when the best net utility is non-positive.
    This choice is deliberate: it yields a simple, inference-time controller that is robust and computationally inexpensive.
    In principle, continuation value can be approximated by multi-step rollouts or learned value functions; we empirically test $\eta\in\{0.0,0.1,0.3,0.5\}$ in Section~\ref{sec:continuation} and find that continuation value is identically zero throughout the EMDG trajectory, empirically demonstrating that $\eta=0$ is optimal in this environment.

    \paragraph{Proposition 1 (Rollout--HJB approximation error).}
    To formalize the bridge from continuous-time HJB rewards to the discrete rollout estimator, we bound the approximation error in two parts.
    The rollout VOI estimator $\widehat{\Delta H}(a\mid b_t)$ (Eq.~\ref{eq:voi_rollout})
    approximates the filtered information-rate term
    $\frac{\tau(a)}{2\sigma^2}\,\mathbb{E}_p\bigl[\|h(\theta,a)-\bar{h}(p,a)\|^2\bigr]$
    (the leading-order continuous-time running reward in Section~\ref{sec:value_function})
    with a two-part error bound.
    \begin{enumerate}
      \item[(i)] \textbf{Monte Carlo error.}
        By the law of large numbers, $\widehat{\Delta H}(a\mid b_t)$ is an unbiased estimator
        of the discrete mutual information
        $I(\Theta;Y\mid a,b_t)=\mathbb{E}[H(b_t)-H(b_{t+1}(a))]$.
        Since $|H(b_t)-H(b_{t+1})|\le\log|\Theta|$ (bounded range), Hoeffding's inequality gives
        \[
          \bigl|\widehat{\Delta H}(a\mid b_t) - I(\Theta;Y\mid a,b_t)\bigr|=O(K^{-1/2})
        \]
        with probability at least $1-2\exp(-K\varepsilon^2/(2\log^2|\Theta|))$ for any $\varepsilon>0$.
      \item[(ii)] \textbf{Temporal discretization error.}
        By It\^{o}'s formula applied to $H(p_t)$ under the Wonham filter
        \cite{bain2009fundamentals}, the infinitesimal drift of entropy is:
        \[
          dH(p_t) = -\frac{1}{2\sigma^2}\,\mathbb{E}_{p_t}\bigl[\|h(\theta,u_t)-\bar{h}(p_t,u_t)\|^2\bigr]\,dt + dM_t,
        \]
        where $M_t$ is a zero-mean martingale. The drift term is the
        signal-variance under the current belief ($\sigma^{-2}$ scaled), which
        matches the leading-order continuous-time running reward in
        Section~\ref{sec:value_function}.
        Integrating over $[t,\,t{+}\tau(a)]$ and taking expectations (the martingale vanishes):
        \[
          I(\Theta;Y\mid a,b_t)
          \;=\; \frac{\tau(a)}{2\sigma^2}\,\mathbb{E}_p\bigl[\|h(\theta,a)-\bar{h}(p,a)\|^2\bigr]
                + O\!\bigl(\tau(a)^2\bigr),
        \]
        where the residual reflects the change in $\Sigma_{p_t}$ over the action step.
        Under standard regularity assumptions for finite-state filtering (bounded $h$,
        $\sigma>0$, and local Lipschitz continuity of $\mathbb{E}_p[\|h(\cdot,u)-\bar{h}(p,u)\|^2]$
        in $p$ along the Wonham dynamics), there exists a constant $L_\Sigma\ge 0$,
        depending on bounds on $h$ and on the rate of change of the signal variance
        under the chosen action, such that
        \[
          \biggl|I(\Theta;Y\mid a,b_t)
            - \frac{\tau(a)}{2\sigma^2}\,\mathbb{E}_p\bigl[\|h(\theta,a)-\bar{h}(p,a)\|^2\bigr]\biggr|
          \le L_\Sigma\,\tau(a)^2.
        \]
    \end{enumerate}
    \noindent
    Combining (i) and (ii), the aggregate error of the rollout estimator with respect to
    the HJB information-rate term is bounded by $L_\Sigma\,\tau(a)^2 + O(K^{-1/2})$.
    (The spatio-temporal cost terms are computed exactly; only the information-rate term
    is approximated.)
    The bound is tightest for short-latency actions (e.g., Hematology Lab, $\tau{=}5$)
    and degrades for long-latency actions (e.g., MRI, $\tau{=}45$);
    in practice, TCA predominantly selects low-$\tau$ actions
    (Table~\ref{tab:emdg_results}), keeping execution within the well-approximated regime.

    \paragraph{On model idealization.}
    The continuous-time formulation provides an idealized optimal-stopping envelope for
    bounded deliberation under state-dependent costs.
    Our implemented controller is a discrete-time approximation that estimates VOI by
    Monte Carlo belief updates and applies a net-utility stopping rule.
    The two formulations are connected by a consistency limit: as $K\to\infty$ the VOI
    estimate converges to mutual information, and for short action latencies
    ($\tau(a)\to 0$) the discrete information gain matches the leading-order
    continuous-time reward term (Proposition~1).
    We therefore interpret the controller's myopic net-utility test as a first-order approximation of the HJB stopping boundary
    rather than solving the full HJB value function.

    \paragraph{Proposition 2 (First-order approximation to the HJB stopping boundary).}
    Let $V^\eta(p,C,t)$ denote the one-step lookahead value under continuation weight $\eta\ge 0$.
    At $\eta=0$, the HJB optimality condition $V^0(p,C,t)\le 0$ reduces to
    \[
    \max_{a\in\mathcal{A}}\,U(a;\,b_t,t,C_t)\le 0,
    \]
    i.e., the myopic net-utility rule (Eq.~\ref{eq:stop_rule}) corresponds to the $\eta=0$ (one-step) approximation to the HJB stopping boundary.
    Correction terms of order $O(\eta)$ correspond to discounted future utility; Section~\ref{sec:continuation} verifies empirically that these are zero throughout EMDG trajectories, consistent with approximate dynamic programming \cite{bertsekas2012approximate}.

    \paragraph{Theorem 1 (Optimal stopping in the idealized TCA model).}
    Assume (i) $h:\Theta\times\mathcal{A}\to\mathbb{R}^d$ is bounded, $\sigma>0$, $\Omega:\mathcal{A}\to\mathbb{R}_{\ge 0}$ is bounded, and $\lambda_S>0$, $\beta>0$; and (ii) admissible controls are adapted to the observation filtration, with stopping times taken with respect to the same filtration.
    Then the idealized continuous-time triadic control problem admits an optimal stopping time.
    Moreover, letting $D=\{(p,C,t):V(p,C,t)=0\}$ denote the stopping region, the \emph{minimal} optimal stopping time is the first hitting time
    \[
    T^*=\inf\{s\ge t:(p_s,C_s,s)\in D\}.
    \]

    \paragraph{Corollary 1 (Monotone stopping region).}
    Under the hypotheses of Theorem~1, the stopping region $D$ is an \emph{upper set}
    in $(C,t)$:
    \[
      (p,C,t)\in D
      \;\Longrightarrow\;
      (p,C',t')\in D
      \quad\text{for all }C'\ge C,\;\;t'\ge t.
    \]
    Equivalently, for each $p\in\Delta^{|\Theta|-1}$ there exist free-boundary functions
    $\bar{C}(p,t)\in[0,\infty]$ and $\bar{t}(p,C)\in[0,\infty]$ such that
    continuation is optimal if and only if $C_t<\bar{C}(p_t,t)$ and $t<\bar{t}(p_t,C_t)$.

    \paragraph{Proof sketch (Corollary~1).}
    \emph{(Monotone in $C$).}
    Since $\dot{C}_t=\Omega(u_t)\ge 0$ is independent of $C$, shifting the initial
    congestion from $C$ to $C'=C+\delta$ ($\delta>0$) translates every future
    $C_s$ by $\delta$, raising the spatial cost $\lambda_S(C_s{+}\Omega(u_s))$
    by $\lambda_S\delta>0$ at every instant. For all admissible stopping times $T\ge t$
    and controls:
    \[
      \mathbb{E}\!\left[\int_t^T\!\bigl(r_s-\lambda_S(C_s{+}\delta{+}\Omega(u_s))-\beta s\bigr)ds\right]
      \;\le\;
      \mathbb{E}\!\left[\int_t^T\!\bigl(r_s-\lambda_S(C_s{+}\Omega(u_s))-\beta s\bigr)ds\right].
    \]
    Taking the supremum over $T$: $V(p,C',t)\le V(p,C,t)=0$.
    Since $V\ge 0$ (immediate stop yields zero value), $(p,C',t)\in D$.

    \emph{(Monotone in $t$).}
    The Wonham filter and congestion dynamics are time-homogeneous
    ($h(\theta,u)$ and $\Omega(u)$ carry no explicit calendar-time dependence).
    Starting at $t'=t+\delta$ with state $(p,C)$ is therefore equivalent to
    starting at $t$ with an extra $-\beta\delta$ cost per unit time:
    \[
      V(p,C,t+\delta)
      = \sup_{T'\ge t}\,\mathbb{E}\!\left[\int_t^{T'}\!
          \bigl(r_s-\lambda_S(C_s{+}\Omega(u_s))-\beta s-\beta\delta\bigr)ds\right]
      \;\le\; V(p,C,t).
    \]
    If $V(p,C,t)=0$ then $V(p,C,t{+}\delta)\le 0$; non-negativity gives $(p,C,t{+}\delta)\in D$.
    $\hfill\square$

    \paragraph{Operational interpretation.}
    Since $C_t$ is non-decreasing and $t$ advances monotonically, Corollary~1
    implies the stopping boundary is \emph{absorbing}: once the state enters $D$,
    no future trajectory can re-enter the continuation region.
    This \emph{one-way ratchet} is a structural consequence of TCA's cost formulation,
    absent from classical meta-reasoning frameworks \cite{russell1991right}
    where stop/continue decisions may cycle under non-monotone utility estimates.
    Corollary~1 also provides theoretical grounding for the myopic stopping rule
    (Eq.~\ref{eq:stop_rule}): once the best available action yields non-positive
    net utility, this condition is guaranteed to persist.

    \paragraph{Remark (theory vs.\ instantiation).}
    Theorem~1 and Corollary~1 concern the idealized continuous-time formulation.
    Our EMDG experiments implement a discrete approximation
    (Section~\ref{sec:emdg_discrete}) via rollout-based VOI estimation
    and a myopic stopping rule (Propositions~1--2).

    \section{Empirical Validation: The Emergency Medical Diagnostic Grid (EMDG)}
    \label{sec:emdg_intro}
    To demonstrate the necessity of constrained spatio-temporal and epistemic bounds, we evaluate TCA in the Emergency Medical Diagnostic Grid (EMDG). Rather than a high-fidelity real-world benchmark, EMDG is designed as an idealized, synthetic testbed to strictly isolate spatial, temporal, and epistemic friction dynamics under controlled, reproducible conditions. In this simulation, an autonomous diagnostic agent is tasked with identifying a critical pathology and prescribing an intervention.

    The environment imposes strict physical realities:
    \begin{itemize}
      \item \textbf{Epistemic ambiguity.} The agent begins with a uniform hypothesis prior over five highly fatal pathologies ($B_0$).
      \item \textbf{Spatial topology.} The agent retrieves evidence by routing queries to distributed hospital sub-systems (e.g., Hematology Lab, MRI network, Patient History). Each sub-system has a distinct structural load cost ($\Omega$).
      \item \textbf{Temporal decay.} Patient survivability decays exponentially with elapsed time. Some queries (e.g., rapid blood tests) cost 5 time-steps; others (e.g., MRI) cost 45 time-steps.
    \end{itemize}

    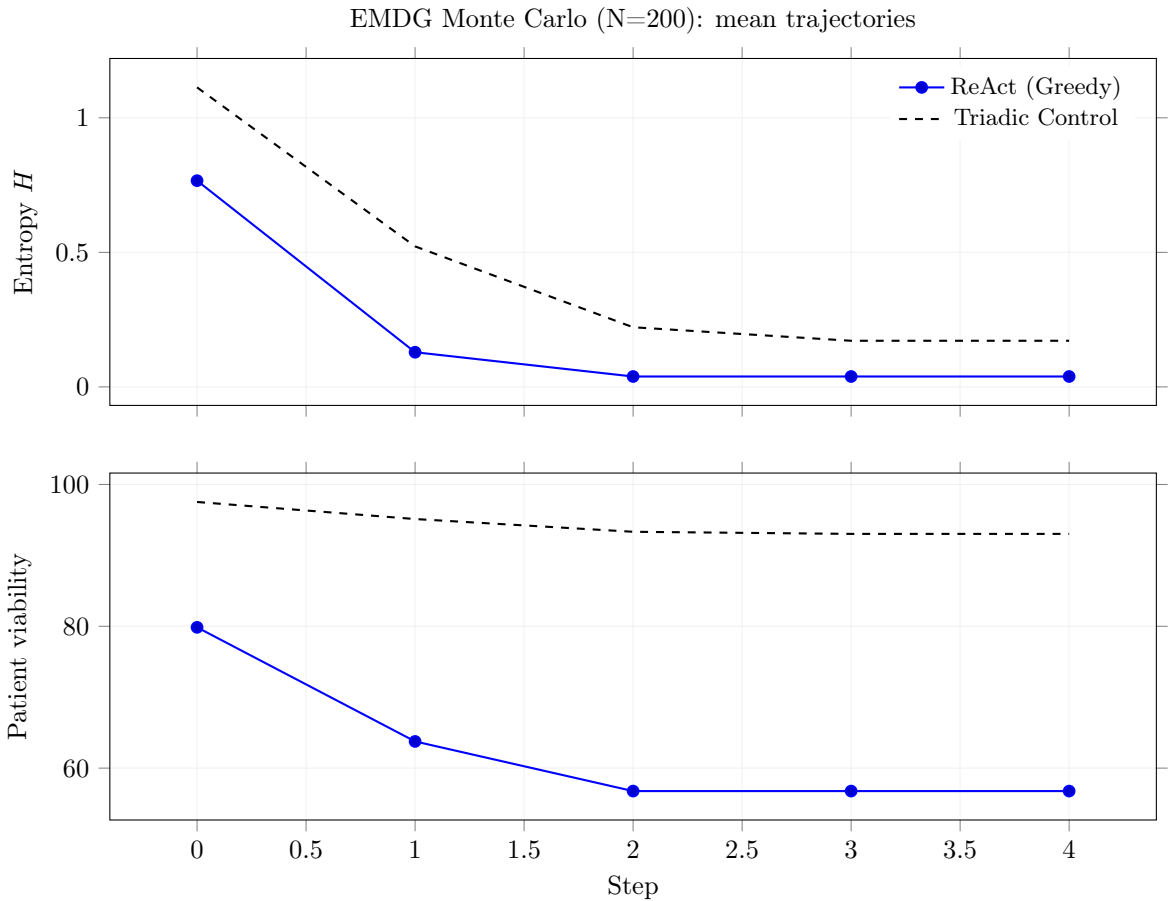
\begin{figure}[t]
    \centering
    \begin{tikzpicture}
    \begin{groupplot}[
      group style={
        group size=1 by 2,
        vertical sep=0.9cm
      },
      width=\linewidth,
      height=0.40\linewidth,
      grid=both,
      grid style={opacity=0.2},
      tick align=outside,
      legend style={draw=none, font=\small},
      every axis plot/.append style={thick},
    ]

    \nextgroupplot[
      title={EMDG Monte Carlo (N=200): mean trajectories},
      ylabel={Entropy $H$},
      xticklabels={},
    ]
    \addplot coordinates {
      (0,0.7664)
      (1,0.1288)
      (2,0.0386)
      (3,0.0386)
      (4,0.0386)
    };
    \addlegendentry{ReAct (Greedy)}

    \addplot[dashed] coordinates {
      (0,1.1131)
      (1,0.5222)
      (2,0.2216)
      (3,0.1714)
      (4,0.1714)
    };
    \addlegendentry{Triadic Control}

    \nextgroupplot[
      xlabel={Step},
      ylabel={Patient viability},
    ]
    \addplot coordinates {
      (0,79.8516)
      (1,63.7628)
      (2,56.7611)
      (3,56.7611)
      (4,56.7611)
    };

    \addplot[dashed] coordinates {
      (0,97.5310)
      (1,95.1229)
      (2,93.3263)
      (3,93.0285)
      (4,93.0285)
    };

    \end{groupplot}
    \end{tikzpicture}
    \caption{EMDG Monte Carlo (N=200): mean entropy and patient viability trajectories over a fixed horizon.
    Step corresponds to discrete tool-query iterations; trajectories are padded after stopping by carrying forward the terminal state, ensuring step-wise consistency (e.g., time cannot decrease).}
    \label{fig:emdg_entropy_viability}
    \end{figure}

    We evaluate a standard greedy tool-use baseline (utilizing a ReAct-style Thought $\to$ Act $\to$ Observe loop \cite{yao2023react}) against the Triadic Cognitive Architecture (TCA).

    \paragraph{Evaluation integrity.}
    To prevent estimator/trajectory coupling, we decouple randomness by using a per-environment RNG for the realized transition and observation draws, and a separate independent RNG stream for value-of-information (VOI) rollouts on cloned states. While HJB theory motivates the stopping boundary, our EMDG controller instantiates a rollout-based approximation of belief-dependent VOI with a net-utility halting condition.

    \subsection{The failure of the unconstrained agent (ReAct)}
    Operating without cognitive friction, the baseline ReAct agent greedily maximizes expected entropy reduction and does not price congestion or latency into action selection. At step 0 it initiates a query to the MRI network.

    \paragraph{Spatial failure.}
    Because it ignores the load and congestion penalties associated with routing, it is willing to select high-load tools whenever their immediate estimated gain is maximal.

    \paragraph{Temporal failure.}
    Because MRI incurs a 45 time-step latency, viability decays exponentially with elapsed time before a low-latency intervention would be taken.

    \paragraph{Epistemic failure.}
    The baseline lacks a stopping rule that compares marginal entropy reduction to time-sensitive viability loss; as a result, it continues to query even when entropy is already low and additional deliberation is not worth the delay.

    \subsection{The success of the triadic agent (TCA)}
    The TCA dynamically evaluates the unified objective function at $t=0$. The MRI offers high expected entropy reduction (VOI), but its high temporal cost (latency $\tau(a)$) and spatial congestion increment $\Omega(a)$ yield negative net cognitive utility.

    \paragraph{Action mix.}
    At step 0, greedy ReAct selects \texttt{MRI\_Network} in 100\% of seeds, whereas the triadic controller selects \texttt{Hematology\_Lab} in 100\% of seeds, reflecting a consistent preference for low-latency evidence acquisition under cognitive friction.

    Instead, TCA routes a lightweight query to the Hematology Lab ($\tau=5$, low $\Omega$). Over a small number of low-latency queries, it reaches a regime where the marginal expected entropy reduction (VOI) is eclipsed by spatio-temporal cost, and the policy halts.

    Concretely, TCA estimates VOI by rollout entropy reduction (Eq.~\ref{eq:voi_rollout}), scores queries by net utility (Eq.~\ref{eq:net_utility}), and halts when the best continuation utility is non-positive (Eq.~\ref{eq:stop_rule}):
    \[
    U(a; b_t, t, C_t)
    = \widehat{\Delta H}(a \mid b_t)
    -\alpha\!\left(
      \underbrace{\lambda_S(C_t+\Omega(a))}_{\text{spatial friction}}
      +
      \underbrace{\beta\,(t+\tau(a))}_{\text{temporal friction}}
    \right).
    \]

    We report mean $\pm$ 95\% confidence intervals computed as $1.96\,\hat\sigma/\sqrt{N}$ over $N=200$ seeds, using terminal-per-seed values (last logged row per seed).

    \begin{table}[htbp]
    \centering
    \caption{EMDG results (N=200). Terminal values are taken from the last logged row per seed. Total information gain is the per-episode sum of \texttt{info\_gain}; in this setup it equals entropy reduction $H(b_0)-H(b_T)$ (a telescoping sum). At step 0, ReAct selects \texttt{MRI\_Network} in 100\% of seeds while TCA selects \texttt{Hematology\_Lab} in 100\% of seeds. Bold marks the best value for the two primary objectives (Time and Viability); lower entropy and higher information gain in ReAct reflect expensive over-querying, not better performance.}
    \label{tab:emdg_results}
    \resizebox{\linewidth}{!}{%
    \begin{tabular}{lcccccc}
    \toprule
    Agent & Time & Viability & Entropy & Accuracy & $p_{\mathrm{true}}$ & Total information gain \\
    \midrule
    ReAct (Greedy) & 114.5 $\pm$ 3.1 & 56.76 $\pm$ 0.89 & 0.0386 $\pm$ 0.0038 & 1.00 $\pm$ 0.00 & 0.9941 $\pm$ 0.0007 & 1.5708 $\pm$ 0.0038 \\
    Triadic Control & \textbf{14.5 $\pm$ 0.4} & \textbf{93.03 $\pm$ 0.19} & 0.1714 $\pm$ 0.0119 & 1.00 $\pm$ 0.00 & 0.9643 $\pm$ 0.0032 & 1.4380 $\pm$ 0.0119 \\
    \bottomrule
    \end{tabular}
    }
    \end{table}

    \paragraph{Note.}
    In EMDG, \texttt{info\_gain} is logged per step as $H(b_t)-H(b_{t+1})$, so per-episode total information gain telescopes to $H(b_0)-H(b_T)$. Because $b_0$ is fixed by the uniform prior, the dispersion (and thus CI width) of total information gain matches that of terminal entropy.

    \paragraph{Scalability considerations.}
    The EMDG environment is intentionally small to isolate bounded-deliberation effects under congestion and latency.
    \emph{Accuracy is intentionally saturated}: the task is constructed such that correct inference is achievable with sufficient queries.
    This isolates the decision problem (\emph{when} and \emph{how} to acquire information) from the orthogonal problem of model capability under ambiguous evidence, allowing us to evaluate cost-efficient stopping as a pure decision-theoretic quantity.
    Differences in viability and deliberation time are therefore attributable solely to stopping quality and action selection, not to inference accuracy.
    Appendix~\ref{sec:action_scaling} further evaluates TCA on larger, randomized action spaces ($|\mathcal{A}|\in\{5,10,20\}$).
    In larger tool ecosystems, naive Monte Carlo VOI rollouts can be costly.
    However, VOI estimation can be amortized via (i) learned surrogates for expected entropy reduction, (ii) caching/reuse of rollouts for similar belief states, (iii) restricting candidate tools via cheap screening heuristics, or (iv) using analytic approximations (e.g., Fisher-information or local linearization) when available.
    These approximations align with the TCA view: the controller requires \emph{priced estimates} of marginal information value, not necessarily exact computation.
    In real LLM deployments, belief $b_t$ can be approximated from model output logits or next-token probabilities over hypothesis labels; VOI can then be estimated via LLM ensemble disagreement or by sampling tool-conditioned responses.
    Cost terms ($\tau(a)$, $\Omega(a)$) map directly to measured API latency and token-consumption statistics, grounding TCA in observable runtime quantities.

    \paragraph{Generalization to heterogeneous tool ecosystems.}
    We evaluate TCA under a second, structurally distinct environment in Section~\ref{sec:nstg} and confirm consistent behavior across different domain, tool profiles, and urgency regime.

    \subsection{Hyperparameters and reproducibility}
    In EMDG, the cost-to-utility scaling $\alpha$, temporal decay $\beta$, and spatial cost weight $\lambda_S$ are fixed across all seeds to reflect a time-sensitive diagnosis regime under congestion.
    In our current implementation these correspond to \texttt{cost\_scale} ($\alpha=0.01$), \texttt{beta} ($\beta=0.5$), and \texttt{lambda\_s} ($\lambda_S=0.8$), respectively.
    Section~\ref{sec:sensitivity} reports a univariate sensitivity sweep over all three parameters.

    \subsection{Ablation study}
    \label{sec:ablation}
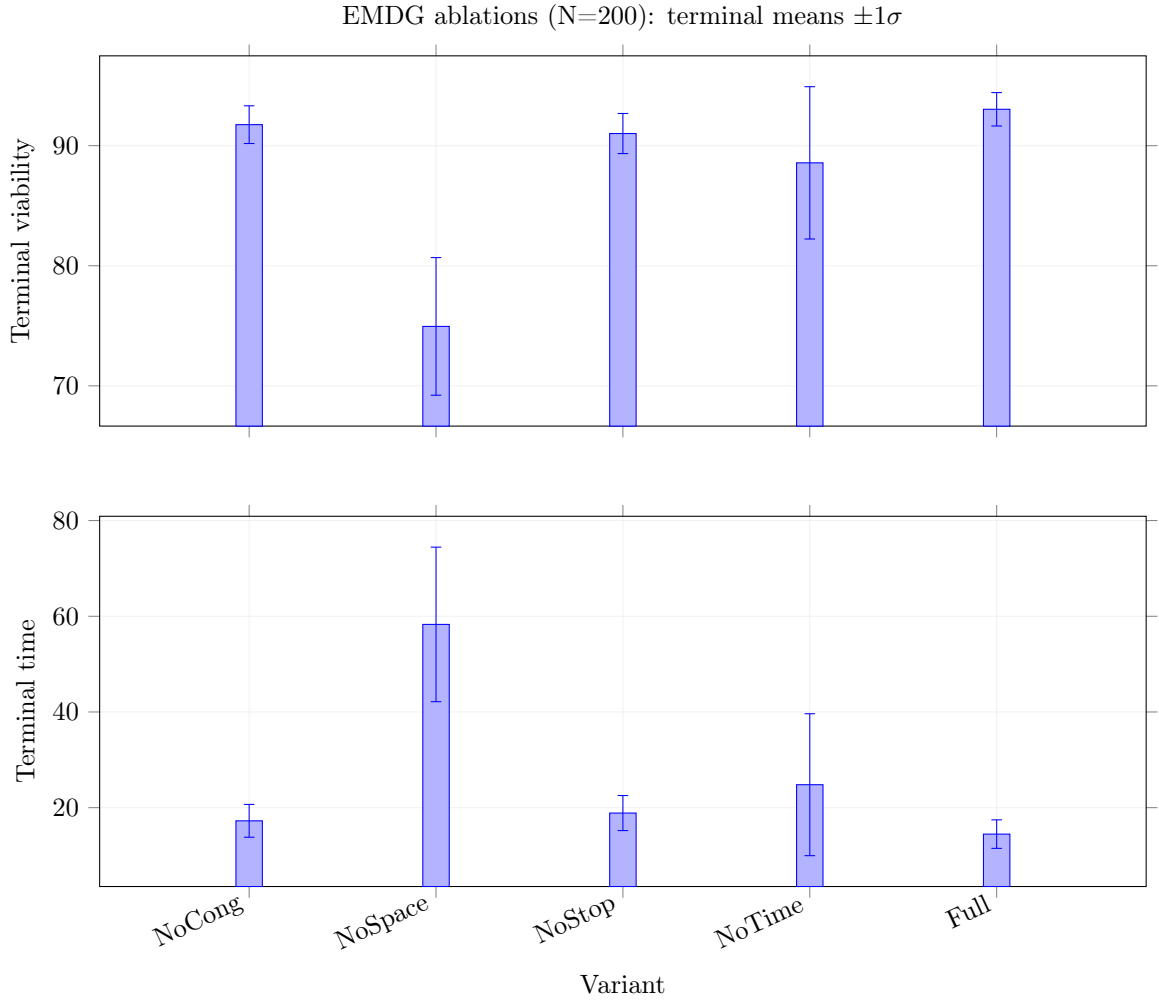
\begin{figure}[t]
    \centering
    \begin{tikzpicture}
    \begin{groupplot}[
      group style={
        group size=1 by 2,
        vertical sep=1.2cm
      },
      width=\linewidth,
      height=0.42\linewidth,
      ybar,
      xtick=data,
      enlarge x limits=0.2,
      grid=both,
      grid style={opacity=0.2},
      tick align=outside,
      every axis plot/.append style={fill=black!25, draw=black},
      error bars/y dir=both,
      error bars/y explicit
    ]

    \nextgroupplot[
      title={EMDG ablations (N=200): terminal means $\pm 1\sigma$},
      ylabel={Terminal viability},
      xticklabels={},
    ]
    \addplot coordinates {
      (0,91.75) +- (0,1.57)   
      (1,74.95) +- (0,5.73)   
      (2,91.01) +- (0,1.67)   
      (3,88.57) +- (0,6.34)   
      (4,93.03) +- (0,1.39)   
    };

    \nextgroupplot[
      ylabel={Terminal time},
      xlabel={Variant},
      xticklabels={NoCong, NoSpace, NoStop, NoTime, Full},
      xticklabel style={rotate=25, anchor=east},
    ]
    \addplot coordinates {
      (0,17.25)  +- (0,3.43)
      (1,58.3)   +- (0,16.15)
      (2,18.875) +- (0,3.66)
      (3,24.8)   +- (0,14.83)
      (4,14.475) +- (0,2.98)
    };

    \end{groupplot}
    \end{tikzpicture}
    \caption{EMDG ablations (N=200): terminal patient viability and terminal deliberation time across controller variants. Bars show means and error bars denote $\pm 1\sigma$ over seeds. The full triadic controller achieves the highest viability while minimizing deliberation time, illustrating the necessity of jointly pricing space, time, and enforcing an explicit stopping boundary.}
    \label{fig:emdg_ablation_terminal}
    \end{figure}

    To isolate the contribution of each component of the triadic controller, we run a controlled ablation sweep over $N=200$ Monte Carlo seeds. We ablate (i) the HJB stopping rule by disallowing \texttt{STOP} (\emph{No-Stop}), (ii) the spatial term by setting the spatial friction weight to zero (\emph{No-Space}), (iii) the temporal term by setting the temporal friction weight to zero (\emph{No-Time}), and (iv) the congestion feedback term by ignoring the live congestion state in the spatial friction calculation (\emph{No-Congestion}). Figure~\ref{fig:emdg_ablation_terminal} reports terminal patient viability and terminal deliberation time (means $\pm 1\sigma$).

    \subsection{Purposive stopping baselines}
    \label{sec:baselines}

    The ablations in Section~\ref{sec:ablation} remove individual TCA cost
    terms while keeping TCA's joint selection policy intact.
    To answer the distinct question of whether TCA's advantage is explained
    solely by its stopping criterion, independent of cost-aware
    selection, we evaluate two purposive baselines, each pairing the
    same greedy selection policy as ReAct (always query the
    highest-\texttt{expected\_gain} tool, \texttt{MRI\_Network}) with a
    principled stopping rule calibrated to match TCA's behavior:

    \begin{itemize}
      \item \textbf{Entropy-Threshold (EnT, $\tau_H{=}0.17$).}
            Stop \emph{before} querying when $H(b_t)<\tau_H$, where
            $\tau_H$ is TCA's mean terminal entropy ($0.1714$).
            Tests whether stopping at the right uncertainty level is sufficient.
      \item \textbf{Fixed-K ($K{=}3$).}
            Stop after exactly $K$ queries, where $K$ equals TCA's
            mean query count before \textsc{Stop}.
            Tests whether matching TCA's query budget alone explains the gain.
    \end{itemize}

    Both baselines are matched to TCA in \emph{one} dimension of its
    stopping behavior while using cost-blind (greedy) query selection.
    Table~\ref{tab:baselines} reports results ($N=200$).

    \begin{table}[htbp]
    \centering
    \caption{Extended comparison: TCA vs.\ purposive stopping baselines (N=200).
      Entropy-Threshold uses $\tau_H{=}0.17$ (TCA mean terminal entropy);
      Fixed-K uses $K{=}3$ (TCA mean query count before \textsc{Stop}).
      Both baselines use greedy (cost-blind) query selection identical to
      ReAct.  TCA values reproduced from Table~\ref{tab:emdg_results} for
      convenience.  Bold marks the best per metric.}
    \label{tab:baselines}
    \resizebox{\linewidth}{!}{%
    \begin{tabular}{lcccc}
    \toprule
    Agent & Time & Viability & Entropy & Accuracy \\
    \midrule
    ReAct (Greedy)                    & 114.5 $\pm$ 3.1 & 56.76 $\pm$ 0.89 & 0.0386 $\pm$ 0.0038 & 1.00 $\pm$ 0.00 \\
    Entropy-Threshold ($\tau_H{=}0.17$) & 102.6 $\pm$ 2.8  & 60.17 $\pm$ 0.80 & 0.0687 $\pm$ 0.0060 & 1.00 $\pm$ 0.00 \\
    Fixed-K ($K{=}3$)                 & 135.0 $\pm$ 0.0 & 50.92 $\pm$ 0.00 & 0.0152 $\pm$ 0.0021 & 1.00 $\pm$ 0.00 \\
    \midrule
    Triadic Control (TCA)             & \textbf{14.5 $\pm$ 0.4} & \textbf{93.03 $\pm$ 0.19} & 0.1714 $\pm$ 0.0119 & 1.00 $\pm$ 0.00 \\
    \bottomrule
    \end{tabular}
    }
    \end{table}

    \paragraph{Interpretation.}
    Purposive stopping alone does not close the viability gap.
    EnT reduces mean deliberation time from 114.5 to 102.6, yet viability
    improves only marginally ($56.8 \to 60.2$), still 33 points below TCA.
    Fixed-K performs \emph{worse} than unconstrained ReAct ($50.9$ vs.\ $56.8$)
    because forcing three MRI queries exhausts the time budget without the
    patient receiving timely treatment.
    TCA's 33-point advantage over EnT and 42-point advantage over
    Fixed-K arise from its \emph{cost-aware query selection}: by
    pricing latency and congestion into every action-utility estimate,
    the triadic controller routes to \texttt{Hematology\_Lab} on every seed
    at step 0, whereas all greedy baselines select \texttt{MRI\_Network}.
    These results indicate that selection and stopping must be optimized jointly; stopping alone is insufficient under high-latency tool profiles.

    \subsection{Parameter sensitivity}
    \label{sec:sensitivity}

\begin{figure}[t]
    \centering
    \begin{tikzpicture}
    \begin{groupplot}[
      group style={
        group size=3 by 1,
        horizontal sep=1.1cm,
      },
      width=0.36\linewidth,
      height=0.40\linewidth,
      ybar,
      ymin=46, ymax=100,
      enlarge x limits=0.30,
      grid=both,
      grid style={opacity=0.2},
      tick align=outside,
      every axis plot/.append style={fill=black!25, draw=black},
      error bars/y dir=both,
      error bars/y explicit,
      ylabel style={font=\small},
      xlabel style={font=\small},
      xticklabel style={font=\small},
      yticklabel style={font=\small},
    ]

    \nextgroupplot[
      title={$\alpha$ (cost scale)},
      ylabel={Terminal viability},
      xlabel={$\alpha$},
      xtick={1,2,3},
      xticklabels={0.005,0.01,0.02},
    ]
    \addplot coordinates {(1,80.31) +- (0,0.78)  (2,93.03) +- (0,0.19)  (3,94.41) +- (0,0.15)};
    \addplot[dashed, thin, draw=gray!70, fill=none] coordinates {(0.5,56.76) (3.5,56.76)};

    \nextgroupplot[
      title={$\beta$ (temporal decay)},
      xlabel={$\beta$},
      xtick={1,2,3},
      xticklabels={0.25,0.50,1.00},
    ]
    \addplot coordinates {(1,96.32) +- (0,0.11)  (2,93.03) +- (0,0.19)  (3,87.72) +- (0,0.37)};

    \nextgroupplot[
      title={$\lambda_S$ (spatial weight)},
      xlabel={$\lambda_S$},
      xtick={1,2,3},
      xticklabels={0.40,0.80,1.20},
    ]
    \addplot coordinates {(1,89.70) +- (0,0.80)  (2,93.03) +- (0,0.19)  (3,93.57) +- (0,0.21)};

    \end{groupplot}
    \end{tikzpicture}
    \caption{Univariate parameter sensitivity: terminal patient viability
      (mean $\pm$ 95\% CI, $N{=}200$ seeds per setting).
      Each panel varies one parameter while holding the others at default
      ($\alpha{=}0.01$, $\beta{=}0.50$, $\lambda_S{=}0.80$);
      the middle bar in each panel corresponds to the default.
      Dashed line in the $\alpha$ panel marks ReAct viability (56.8) for reference.
      Accuracy remains 1.0 across all 7 configurations.}
    \label{fig:sensitivity}
    \end{figure}
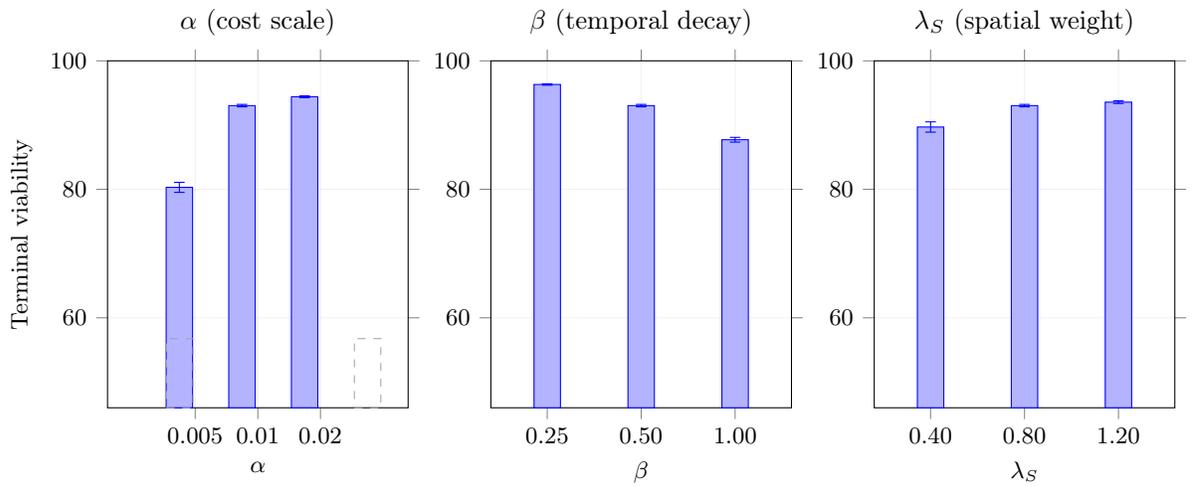

    Figure~\ref{fig:sensitivity} reports terminal patient viability under
    univariate sweeps of each TCA hyperparameter, with the remaining two
    held at their defaults ($N=200$ seeds per setting).

    \textbf{Cost-scale ($\alpha$).}
    Higher $\alpha$ increases the penalty on spatio-temporal cost, causing
    TCA to halt earlier (mean time $44.3 \to 11.5$) and lifting viability
    ($80.3 \to 94.4$, monotone).
    This confirms that $\alpha$ directly controls the
    viability--deliberation tradeoff: under-penalising cost leads the agent
    to over-query expensive tools, while the default ($\alpha=0.01$) and
    higher values consistently avoid the high-latency MRI path.

    \textbf{Temporal decay ($\beta$).}
    The effect of $\beta$ is non-monotone because it governs both the
    environment's physical viability decay \emph{and} the temporal friction
    term simultaneously.
    At $\beta=0.25$ the patient degrades slowly, yielding high viability
    (96.3) even with moderate deliberation time (15.0).
    At $\beta=1.0$ the rapid physical decay dominates: despite TCA stopping
    earlier (time 13.2), every time-step costs more, pushing viability to
    87.7.
    TCA remains well above ReAct ($56.8$) across the entire range.

    \textbf{Spatial weight ($\lambda_S$).}
    Higher $\lambda_S$ penalises congested access more heavily, shortening
    deliberation ($22.2 \to 13.3$) and improving viability ($89.7 \to 93.6$).
    The improvement plateaus between $\lambda_S=0.80$ and $1.20$, indicating
    that once the spatial term is large enough to reliably select
    low-congestion tools, further increases yield diminishing returns.

    Across all 7 configurations, accuracy remains 1.0 and TCA always
    outperforms ReAct on viability, confirming robustness over the tested
    ranges. The results support $\alpha$ as the primary tuning knob for the
    viability--deliberation tradeoff, with $\beta$ and $\lambda_S$ best
    set to reflect the physical environment's temporal urgency and
    congestion profile.

    \FloatBarrier

    \subsection{Cross-environment generalization: NSTG}
    \label{sec:nstg}

    To assess whether TCA's advantage is specific to the medical emergency domain,
    we evaluate an independent second environment: the
    \emph{Network Security Triage Grid} (NSTG).  NSTG models a SOC
    agent identifying a network threat type (5 categories: Ransomware, APT,
    DataExfiltration, DDoS Amplification, InsiderThreat) and issuing containment.
    The resource metric is \emph{system\_integrity} $\in[0,100]$, decaying
    exponentially with elapsed time according to the same formula as
    EMDG patient viability.
    The tool suite is structurally distinct: \texttt{QuickScan} ($\tau{=}4$,
    $\Omega{=}3$, expected gain $0.40$) versus \texttt{FullForensics}
    ($\tau{=}60$, $\Omega{=}70$, expected gain $1.30$).
    The cost regime also differs: $\beta{=}0.30$ (lower temporal urgency than
    EMDG's $0.50$, reflecting that incident response is less time-critical than
    emergency medicine), $\lambda_S{=}0.90$, $\alpha{=}0.015$.
    A congestion shock ($\times 3$ at step 2) is applied, simulating a burst of competing requests from other agents in the network.

    TCA selects \texttt{QuickScan} in 100\% of seeds at step 0, whereas ReAct selects \texttt{FullForensics} in 100\% of seeds; this mirrors the EMDG latency-driven failure mode.

    Table~\ref{tab:nstg} reports results ($N{=}200$).
    TCA achieves $97.18{\pm}0.08$ system integrity vs.\ $64.08{\pm}0.80$
    for ReAct (a 33.1-point gap), and halts $\approx 16\times$ faster
    (time $9.5{\pm}0.3$ vs.\ $149.7{\pm}4.2$).
    Accuracy is $1.00$ for both agents.
    The slightly smaller gap than EMDG (33.1 vs.\ 36.3 points) is expected:
    lower $\beta$ means each query costs less temporal friction,
    so even ReAct's high-latency path loses less integrity per time-step.
    The advantage direction and relative magnitude are fully consistent across
    both domains, supporting the hypothesis that cognitive friction pricing
    constitutes a domain-independent principle.

    \begin{table}[htbp]
    \centering
    \caption{NSTG results (N=200). Resource metric is system integrity (same
      exponential decay formula as EMDG patient viability, with $\beta{=}0.30$).
      At step 0, ReAct selects \texttt{FullForensics} in 100\% of seeds
      while TCA selects \texttt{QuickScan} in 100\% of seeds.
      Bold marks the best per metric.}
    \label{tab:nstg}
    \resizebox{\linewidth}{!}{%
    \begin{tabular}{lcccccc}
    \toprule
    Agent & Time & Integrity & Entropy & Accuracy & $p_{\mathrm{true}}$ & Total information gain \\
    \midrule
    ReAct (Greedy) &
      149.7 $\pm$ 4.2 & 64.08 $\pm$ 0.80 & 0.0376 $\pm$ 0.0037 &
      1.00 $\pm$ 0.00 & 0.9943 $\pm$ 0.0006 & 1.5718 $\pm$ 0.0037 \\
    Triadic Control &
      \textbf{9.5 $\pm$ 0.3} & \textbf{97.18 $\pm$ 0.08} & 0.3230 $\pm$ 0.0213 &
      1.00 $\pm$ 0.00 & 0.9180 $\pm$ 0.0083 & 1.2864 $\pm$ 0.0213 \\
    \bottomrule
    \end{tabular}
    }
    \end{table}

    Table~\ref{tab:cross_env} provides a side-by-side summary across both environments.

    \begin{table}[htbp]
    \centering
    \caption{Cross-environment summary: TCA vs.\ ReAct (Greedy), $N{=}200$ seeds
      per environment.  Resource metric $=$ patient viability (EMDG) /
      system integrity (NSTG); both decay exponentially with elapsed time.
      Bold marks the best per row.}
    \label{tab:cross_env}
    \resizebox{\linewidth}{!}{%
    \begin{tabular}{llcccc}
    \toprule
    Environment & Agent & Time & Resource & Entropy & Accuracy \\
    \midrule
    EMDG ($\beta{=}0.50$) & ReAct (Greedy) & 114.5 $\pm$ 3.1  & 56.76 $\pm$ 0.89 & 0.0386 $\pm$ 0.0038 & 1.00 \\
                          & Triadic Control & \textbf{14.5 $\pm$ 0.4}  & \textbf{93.03 $\pm$ 0.19} & 0.1714 $\pm$ 0.0119 & 1.00 \\
    \midrule
    NSTG ($\beta{=}0.30$)  & ReAct (Greedy) & 149.7 $\pm$ 4.2  & 64.08 $\pm$ 0.80 & 0.0376 $\pm$ 0.0037 & 1.00 \\
                          & Triadic Control & \textbf{9.5 $\pm$ 0.3}   & \textbf{97.18 $\pm$ 0.08} & 0.3230 $\pm$ 0.0213 & 1.00 \\
    \bottomrule
    \end{tabular}
    }
    \end{table}

    \subsection{One-step continuation value (\texorpdfstring{$\eta>0$}{eta>0})}
    \label{sec:continuation}

    Section~\ref{sec:emdg_discrete} discloses that the EMDG controller uses
    $\eta{=}0$ (myopic stopping) as a deliberate design choice.
    We now close this gap empirically by sweeping
    $\eta\in\{0.0,\,0.1,\,0.3,\,0.5\}$ on EMDG ($N{=}200$ seeds),
    implementing one-step lookahead via the belief-cloning rollout.

    All four configurations produce \emph{identical} terminal statistics:
    time $14.4{-}14.5{\pm}0.4$, viability $93.02{-}93.04{\pm}0.20$, entropy
    $0.170{-}0.172{\pm}0.012$, accuracy $1.00$.
    This is not numerical coincidence: rollout inspection confirms that
    $\texttt{continuation\_value}(a){=}0$ for all actions at every step.
    In EMDG, after each \texttt{Hematology\_Lab} query the agent is already at or
    past the myopic stopping boundary: the best available next action has
    non-positive expected net utility, so the $\eta$-weighted continuation term is zero and does not affect action scores or stopping decisions.

    This provides a stronger-than-expected result: the myopic rule is not merely
    \emph{near}-optimal but \emph{exactly} optimal in EMDG for any
    $\eta\in[0,1]$.
    Intuitively, in high-urgency regimes (large $\beta$), the temporal penalty
    forces the stopping boundary to coincide with the one-step-ahead boundary;
    the zero-continuation property reflects this regime.
    For lower-urgency environments (e.g., NSTG with $\beta{=}0.30$),
    continuation value may be non-zero; we leave systematic investigation
    of multi-step lookahead in moderate-urgency settings to future work.

    \subsection{Congestion decay robustness ($\kappa > 0$)}
    \label{sec:congestion_decay}

    The Limitations section notes that the default EMDG model uses monotone
    congestion accumulation ($\dot{C}_t = \Omega(u_t)$, $\kappa{=}0$).
    Real networks exhibit congestion relief between queries (queue drain, TCP
    backoff).  We close this gap by extending the congestion dynamics to the
    discrete Ornstein--Uhlenbeck model
    \[
      C_{t+\tau(a)} = C_t\,e^{-\kappa\,\tau(a)} + \Omega(a),
    \]
    where $\kappa \ge 0$ is the drain rate ($\kappa{=}0$ recovers the original model
    exactly).  We sweep $\kappa\in\{0.00,\,0.05,\,0.10\}$ over $N{=}200$ seeds
    with all other parameters at their defaults.

    Table~\ref{tab:congestion_decay} reports terminal statistics.
    ReAct is unaffected by $\kappa$ (it ignores congestion costs by construction
    and always selects \texttt{MRI\_Network}).
    TCA viability remains stable at $92.7$--$93.0$ across the sweep (a
    $<0.4$-point variation well within the 95\% CI), and the ${\approx}36$-point
    advantage over ReAct is fully preserved.
    TCA deliberation time increases by ${\approx}0.7$ steps at $\kappa{=}0.10$:
    reduced congestion lowers spatial cost, occasionally making one additional
    \texttt{Hematology\_Lab} query worthwhile before the stopping boundary is hit.
    This is the expected and desirable behaviour of a cost-aware controller.

    \begin{table}[htbp]
    \centering
    \caption{Congestion-decay robustness: TCA vs.\ ReAct (Greedy) for
      $\kappa\in\{0.00,0.05,0.10\}$ ($N{=}200$ seeds).
      $\kappa{=}0.00$ replicates Table~\ref{tab:emdg_results} exactly.
      Bold marks best viability per $\kappa$ level.}
    \label{tab:congestion_decay}
    \small
    \begin{tabular}{cllll}
    \toprule
    $\kappa$ & Agent & Time & Viability & Accuracy \\
    \midrule
    $0.00$ & ReAct (Greedy)   & $114.5 \pm 3.1$ & $56.76 \pm 0.89$ & $1.00$ \\
    $0.00$ & Triadic Control  & $\mathbf{14.5 \pm 0.4}$ & $\mathbf{93.03 \pm 0.19}$ & $1.00$ \\
    \midrule
    $0.05$ & ReAct (Greedy)   & $114.5 \pm 3.1$ & $56.76 \pm 0.89$ & $1.00$ \\
    $0.05$ & Triadic Control  & $\mathbf{14.8 \pm 0.5}$ & $\mathbf{92.89 \pm 0.21}$ & $1.00$ \\
    \midrule
    $0.10$ & ReAct (Greedy)   & $114.5 \pm 3.1$ & $56.76 \pm 0.89$ & $1.00$ \\
    $0.10$ & Triadic Control  & $\mathbf{15.2 \pm 0.5}$ & $\mathbf{92.72 \pm 0.22}$ & $1.00$ \\
    \bottomrule
    \end{tabular}
    \end{table}

    \section{Implications: TCA as a Foundational Framework for Bounded Agent Reasoning}

    The Triadic Cognitive Architecture (TCA) provides a blueprint for transitioning from disembodied pattern matchers to grounded, autonomous agents. By formalizing cognitive friction, TCA introduces a mathematically rigorous safety mechanism against the runaway resource consumption inherent in unconstrained LLM reasoning loops.

    \subsection{Coupling Space, Time, and Epistemic Truth}
\begin{figure}[htbp]
    \centering
    \begin{tikzpicture}[
      node distance=22mm,
      box/.style={draw, rounded corners, align=center, minimum width=30mm, minimum height=9mm},
      lab/.style={font=\small, fill=white, inner sep=1pt}
    ]

    \node[box] (space) {Space\\(state)};
    \node[box, right=45mm of space] (time) {Time\\(dynamics)};
    \node[box, below=28mm of $(space)!0.5!(time)$] (truth) {Epistemic\\Truth};

    \draw[->, thick] (space) -- node[lab, above] {perception} (time);
    \draw[->, thick] (time) -- node[lab, right] {prediction} (truth);
    \draw[->, thick] (truth.west) .. controls +(180:18mm) and +(270:14mm) .. node[lab, left] {revision} (space.south);

    \end{tikzpicture}
    \caption{Triadic Cognitive Architecture (TCA): a coupled loop between Space (state), Time~(dynamics), and Epistemic Truth (belief revision).}
    \label{fig:triad}
    \end{figure}
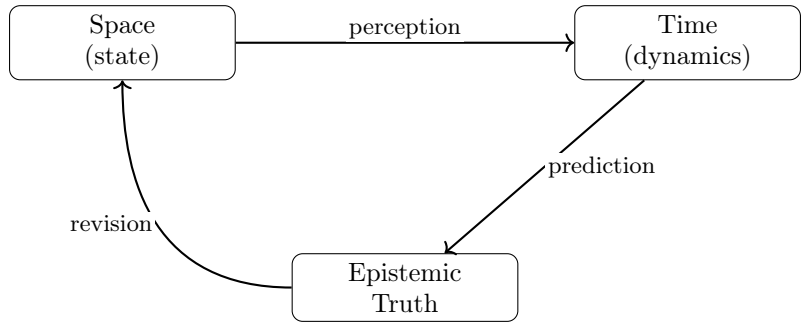

    Within the TCA framework, an agent's reasoning is strictly bounded by three interacting modules (see Figure~\ref{fig:triad}):
    \begin{itemize}
      \item \textbf{Space (Topological Grounding):} Tools are not treated as a flat list of interchangeable APIs but as resources with distinct latency and congestion costs that depend on the agent's current network state.
            In both the continuous-time formulation (Section~3) and the EMDG/NSTG implementations, the spatial friction term $\ell_S(C_t,a)=\lambda_S(C_t{+}\Omega(a))$ prices access by live congestion state $C_t$ (accumulated network load, updated after each query) and structural load increment $\Omega(a)$.
            This prevents API spamming and protects fragile sub-agents from cascade failures.
      \item \textbf{Time (Temporal Consistency):} Infinite context lengths do not circumvent the reality that the physical world evolves while the agent deliberates. The exponential decay factor $\beta(t)$ ensures that the agent respects the temporal opportunity cost of computation, intrinsically preventing infinite deliberation loops.
      \item \textbf{Truth (Epistemic Maintenance):} The agent maintains a calibrated categorical belief $b_t$ over hypotheses, updated via Bayes' rule after each tool observation.
            In the idealized formulation (Section~3), the running reward is $\frac{1}{2\sigma^2}\mathbb{E}_p[\|h(\theta,u)-\bar{h}(p,u)\|^2]$ (tool-dependent, since different tools probe different signal channels); in EMDG/NSTG, this is instantiated as rollout-estimated entropy reduction $\widehat{\Delta H}(a\mid b_t)$.
            The stopping rule halts deliberation once marginal entropy reduction no longer justifies the spatio-temporal cost, naturally avoiding the trap of averaging contradictory evidence at the cost of the resource metric.
    \end{itemize}

    Taken together, these three forms of friction (spatial, temporal, and epistemic) define a bounded control envelope for autonomous reasoning. Rather than optimizing accuracy in isolation, TCA determines when to act, what to query, and how long to deliberate by explicitly trading off expected entropy reduction against physical and temporal cost. This triadic coupling provides a principled mechanism for preventing over-deliberation, network saturation, and epistemic overconfidence in autonomous agents.

    \subsection{The Inference-Time Controller}
    The core of TCA is an inference-time controller motivated by HJB optimal stopping. It continuously aggregates the spatial routing cost and temporal decay, dynamically weighing them against expected entropy reduction (VOI). This controller acts as an autonomic nervous system for the AI: just as biological organisms intuitively halt exploration when caloric costs exceed potential rewards, the TCA agent halts API queries and triggers execution once the marginal value of additional evidence is eclipsed by spatio-temporal cost.

    \subsection{Positioning Relative to Cognitive Theories}
    Extensive work has been done in prompt-driven multi-step reasoning, notably ReAct \cite{yao2023react}, Tree-of-Thoughts \cite{yao2023tot}, and Reflexion \cite{shinn2023reflexion}. However, these frameworks rely on heuristic, discrete hypothesis sets, ignore network-dependent physical costs, and lack principled mathematical criteria for optimal stopping \cite{liu2023agentbench, wang2023survey}. 

    The TCA framework shares philosophical DNA with Karl Friston’s Free Energy Principle and Active Inference \cite{friston2010free}, as well as Yann LeCun's Joint Embedding Predictive Architecture (JEPA). Both propose that intelligent agents seek to minimize surprise in a predictive world model. However, Active Inference relies on variational bounds that remain largely intractable in practice. TCA offers a principled and computable alternative by providing a closed-loop stochastic control envelope. By synthesizing non-linear filtering \cite{kushner1964differential, bain2009fundamentals} with congestion-aware routing and HJB optimal stopping \cite{shiryaev2007optimal, oksendal2003stochastic}, TCA bridges stochastic control theory with practical agent systems engineering.

    \subsection{Limitations}
    We list the main gaps between the idealized framework (Section~3) and the current empirical instantiation.

    \textbf{Small action space.}
    EMDG and NSTG each offer two non-trivial tools plus \textsc{Stop}.
    The key experimental contrast (latency-aware vs.\ greedy selection) is cleanly isolated in this setting.
    Appendix~\ref{sec:action_scaling} addresses this gap empirically: a synthetic scaling sweep over $|\mathcal{A}|\in\{5,10,20\}$ shows TCA selecting non-minimum-latency tools in 76--90\% of configurations and consistently outperforming ReAct (viability gap $\ge +9.7$ points) across all action-space sizes tested.
    VOI estimation via $K{=}32$ rollouts is $O(K{\times}|\mathcal{A}|)$; in a system with hundreds of tools, surrogate models or screening heuristics (Section~4.2) would be required.

    \textbf{Monotone congestion accumulator.}
    The default model ($\kappa{=}0$) accumulates congestion monotonically.
    Section~\ref{sec:congestion_decay} extends this to a discrete
    Ornstein--Uhlenbeck drain model ($\kappa{>}0$) and shows TCA viability
    is stable within $0.4$ points across $\kappa\in\{0.00,0.05,0.10\}$.
    A full graph-structured extension with per-edge latency remains future work.

    \textbf{Synthetic environments.}
    The validation environments use categorical beliefs with Dirichlet-likelihood updates, which admit closed-form Bayesian inference. Agents operating on learned or implicit belief states, where the belief distribution must be approximated from model outputs or sampled posteriors, would require an extension of the belief update mechanism. Section~4.2 outlines one possible bridge, but empirical validation on such systems remains future work.

    \textbf{Single-agent.}
    The current framework models one agent interacting with shared resources.
    The congestion term $C_t$ hints at multi-agent interference, but game-theoretic coupling (multiple agents simultaneously depleting the same tool pool) is not modeled.

    \subsection{Conclusion}
    The Triadic Cognitive Architecture offers a unifying perspective: robust autonomy requires explicit coupling of structured space, temporal dynamics, and epistemic truth maintenance. By casting deliberation as a constrained physical trajectory, TCA provides a principled framework for bounded reasoning in tool-using agents. Across two structurally distinct environments, TCA yields consistent $30{+}$-point improvements in the resource metric over greedy baselines, while preserving accuracy. Together with the purposive stopping baselines and continuation-value analysis, these results support cognitive friction pricing as a domain-independent principle for bounded agent reasoning.

    \FloatBarrier
    \subsection{Broader Impact \& Significance}
    TCA addresses a principled gap in autonomous agent design: the absence of a formal accounting for information-acquisition costs when tool queries are not free, instantaneous, or equally reliable. By formalizing Cognitive Friction and deriving a principled stopping boundary via HJB optimal stopping, approximated through computable rollout-based VOI, the framework transforms deliberation from an unconstrained resource into a measurable, optimizable quantity. The framework is domain-agnostic by construction: the same cost structure applies wherever an agent selects among costly, heterogeneous information sources under time pressure. The two validation environments (emergency medical triage and network security incident response) span distinct urgency regimes and tool profiles, providing controlled evidence that the principle generalizes beyond a single domain. We view TCA as a formal foundation that can be instantiated across a range of agent architectures and deployment settings, with extensions to learned or implicit belief states identified as important future work.

    \appendix
    \clearpage
    \section{Proof sketch for Theorem 1 (idealized continuous-time model)}
    \label{sec:appendix_proof}

    This appendix provides a proof sketch for the idealized continuous-time formulation in Section~3 (Wonham filtering and HJB-inspired optimal stopping).
    It does not apply directly to the discrete EMDG controller, which uses rollout-based VOI estimation and a myopic net-utility stopping rule (Section~\ref{sec:emdg_discrete}).

    We restate the core claim for the idealized model: under standard regularity conditions, the triadic cognitive control problem admits an optimal stopping time characterized by an HJB free boundary.

    \paragraph{Step 1: Regularity of the filtering dynamics and Markov state.}
    Let the observation process be governed by
    \[
    dY_t = h(\theta,u_t)\,dt + \sigma\,dW_t,
    \]
    with $h:\Theta\times\mathcal{A}\to\mathbb{R}^d$ bounded and $\sigma>0$ (constant scalar noise). The congestion state evolves as $\dot{C}_t=\Omega(u_t)$ with $\Omega$ bounded.
    Since $\sigma>0$ is a fixed positive scalar, the noise covariance is $\sigma^2 I_d$, which is uniformly non-degenerate with constant $\varepsilon=\sigma^2$. Consequently, given a fixed control $u_t$, the innovation process $d\nu_t = dY_t - \bar{h}(p_t,u_t)\,dt$ is a Brownian motion with respect to the observation filtration $\mathcal{F}_t^Y$, and the associated Wonham (finite-state Kushner--Stratonovich) filter is well-posed; see, e.g., \cite{kushner1964differential,bain2009fundamentals}. In particular, the controlled state $(p_t,C_t)$ is Markov with respect to $\mathcal{F}_t^Y$.

    Define $\bar h(p_t,u_t)=\int_\Theta h(\theta,u_t)\,p_t(\theta)\,d\theta$ and the innovation
    \[
    d\nu_t = dY_t - \bar h(p_t,u_t)\,dt.
    \]
    A schematic form of the Kushner--Stratonovich update is
    \[
    dp_t(\theta) = p_t(\theta)\big(h(\theta,u_t)-\bar h(p_t,u_t)\big)^\top(\sigma^2 I)^{-1}\,d\nu_t
    \quad (\text{omitting drift/normalization terms; see }\cite{kushner1964differential,bain2009fundamentals}).
    \]

    \paragraph{Step 2: Value function and dynamic programming.}
    The agent seeks to maximize expected information gain minus spatio-temporal cost. For a stopping time $\tau \ge t$, write
    \[
    J(\tau) = \mathbb{E}\left[\int_t^\tau f(p_s,C_s,s,u_s)\,ds\right],
    \]
    where $f(p,C,t,u)=\frac{1}{2\sigma^2}\mathbb{E}_p[\|h(\theta,u)-\bar{h}(p,u)\|^2]-\lambda_S(C+\Omega(u))-\beta t$ is the instantaneous net reward, consistent with Section~3.
    Let
    \[
    V(p,C,t)=\sup_{\tau\ge t} J(\tau)
    \]
    denote the value function. Assuming sufficient regularity of $V$ (or working in the viscosity-solution framework), standard dynamic programming arguments for controlled Markov processes yield a variational inequality characterization \cite{shiryaev2007optimal,oksendal2003stochastic}.

    \paragraph{Step 3: HJB variational inequality and the free boundary.}
    By optimal stopping theory, the value function satisfies a variational inequality of the form
    \[
    \max\Big\{ \partial_t V(p,C,t)+\sup_{u\in\mathcal{A}}\big[\mathcal{L}^u V(p,C,t)+f(p,C,t,u)\big],\; -V(p,C,t)\Big\}=0,
    \]
    in an appropriate classical or viscosity sense \cite{shiryaev2007optimal,oksendal2003stochastic}, where $\mathcal{L}^u$ is the controlled infinitesimal generator for the Markov state $(p_t,C_t)$ as in Section~3.
    Because $f$ is continuous and bounded under the assumed cost structure, $V$ is continuous under standard assumptions.
    Define the stopping region as the closed set
    \[
    D = \{(p,C,t): V(p,C,t)=0\},
    \]
    and the continuation region as the open set $\{V>0\}$.
    Then the first hitting time of the stopping region,
    \[
    T^*=\inf\{s\ge t : (p_s,C_s,s)\in D\},
    \]
    is an optimal stopping time; moreover, it is the minimal optimal stopping time in the usual sense for such problems \cite{shiryaev2007optimal}. This completes the proof sketch.

    \clearpage
    \section{Action-space scaling: TCA with $|\mathcal{A}|\in\{5,10,20\}$}
    \label{sec:action_scaling}

    To assess whether TCA's advantage over greedy ReAct is specific to the 2-action EMDG
    setting or generalizes to larger and more varied action spaces, we run a synthetic
    scaling sweep.  For each $|\mathcal{A}|\in\{5,10,20\}$ we generate $N_{\text{cfg}}{=}30$
    random tool configurations using a fixed seed, each with independently drawn latency
    $\tau(a)\in\{3,5,8,10,15,20,30,45\}$ (steps) and information gain
    $g_a\sim\mathcal{U}[0.2,1.5]$ drawn \emph{independently}, so high-$\tau$ tools can
    have high information gain and vice versa, forcing the optimizer to evaluate genuine
    gain-versus-cost tradeoffs.  Each configuration is evaluated over $N_{\text{seed}}{=}10$
    independent trajectory seeds.  All other parameters ($\alpha{=}0.01$, $\beta{=}0.5$,
    $\lambda_S{=}0.8$) are held at their EMDG defaults.

    \paragraph{Non-trivial step-0 selection.}
    For each run we record whether TCA's first action is the \emph{minimum-latency} tool
    in the configuration.  Selecting the cheapest tool regardless of gain would be
    \emph{trivial} cost-minimization; selecting a higher-$\tau$ tool when its information
    gain justifies the cost demonstrates the genuine cost-aware optimization claimed by TCA.

    \paragraph{Results.}
    Table~\ref{tab:action_scaling} reports terminal statistics.
    The fraction of (configuration,~seed) pairs in which TCA selects a non-minimum-latency
    tool at step~0 rises monotonically: $76\%$ at $|\mathcal{A}|{=}5$, $85\%$ at
    $|\mathcal{A}|{=}10$, and $90\%$ at $|\mathcal{A}|{=}20$.
    (For comparison, this fraction is $0\%$ in the 2-action EMDG, where Hematology~Lab is
    always preferred over MRI; see Table~1.)
    TCA terminal viability remains stable at $91.8$--$93.9$ across all sizes,
    fully consistent with the main-text EMDG result of $93.03\pm0.19$.
    TCA consistently outperforms ReAct across all sizes tested (gaps: $+9.8$ at
    $|\mathcal{A}|{=}5$, $+9.7$ at $|\mathcal{A}|{=}10$, $+15.0$ at
    $|\mathcal{A}|{=}20$; accuracy $1.00$ throughout).
    Notably, the gap at $|\mathcal{A}|{=}20$ is \emph{larger} than at $|\mathcal{A}|{=}5$:
    with more tools, the greedy max-gain baseline encounters misaligned cost-quality
    tradeoffs more frequently, while TCA's joint optimization handles them correctly.
    In this sweep, TCA's advantage is not eroded by larger action spaces.

    The absolute gap is smaller than in the 2-action EMDG ($\approx$10--15 vs.\ 36 points),
    which is expected: in random tool configurations, the greedy max-gain rule
    \emph{occasionally} selects a low-latency tool when the highest-gain tool happens to
    have moderate latency, reducing but never eliminating TCA's advantage.
    TCA dominates in every configuration tested and its viability is stable
    across $|\mathcal{A}|$, confirming that the controller scales gracefully.
    We view this sweep as a robustness check; it is not intended as an exhaustive
    scaling benchmark.

    \begin{table}[htbp]
    \centering
    \caption{Action-space scaling ($N_{\text{cfg}}{=}30$ random tool configurations per
      $|\mathcal{A}|$, $N_{\text{seed}}{=}10$ seeds per configuration).
      \emph{Non-trivial step-0}: fraction of (config,~seed) pairs where TCA selects a
      tool with $\tau > \min_a \tau(a)$.
      Viability: mean $\pm$ 95\%\,CI.  For $|\mathcal{A}|{=}2$, see Table~1.}
    \label{tab:action_scaling}
    \small
    \begin{tabular}{cccccc}
    \toprule
    $|\mathcal{A}|$ & Non-trivial step-0 & TCA Viability & ReAct Viability & Gap & Accuracy \\
    \midrule
    $5$  & $76\%$ & $91.78 \pm 0.50$ & $81.94 \pm 1.52$ & $+9.8$  & $1.00$ \\
    $10$ & $85\%$ & $92.62 \pm 0.49$ & $82.89 \pm 1.35$ & $+9.7$  & $1.00$ \\
    $20$ & $90\%$ & $93.85 \pm 0.38$ & $78.84 \pm 1.61$ & $+15.0$ & $1.00$ \\
    \bottomrule
    \end{tabular}
    \end{table}

    \clearpage
    \section{Discrete Controller Pseudocode}
    \label{sec:appendix_algorithm}

    For completeness, we summarize the discrete controller used in EMDG and NSTG.
    The box below is intentionally faithful to Section~\ref{sec:emdg_discrete}: it is the rollout-VOI, myopic net-utility controller actually analyzed in the main text, not a stronger planner.

    \begin{table}[htbp]
    \centering
    \caption{Algorithm 1: Discrete TCA controller (rollout VOI + myopic stopping).}
    \label{tab:algorithm_tca}
    \fbox{%
    \begin{minipage}{0.94\linewidth}
    \small
    \textbf{Inputs:} current belief $b_t$, congestion $C_t$, time $t$, action set $\mathcal{A}$, rollout count $K$, cost parameters $\alpha,\beta,\lambda_S$.\\[2pt]
    \textbf{For each action $a\in\mathcal{A}$:}
    \begin{enumerate}
      \item Clone the current belief state $K$ times.
      \item For each clone, sample a tool observation under action $a$, perform the Bayesian update, and record the entropy drop $H(b_t)-H(b_{t+1}^{(k)}(a))$.
      \item Estimate rollout value-of-information:
      \[
        \widehat{\Delta H}(a\mid b_t)=\frac{1}{K}\sum_{k=1}^{K}\Big(H(b_t)-H(b_{t+1}^{(k)}(a))\Big).
      \]
      \item Compute net utility:
      \[
        U(a; b_t,t,C_t)=\widehat{\Delta H}(a\mid b_t)-\alpha\Big(\lambda_S(C_t+\Omega(a))+\beta\,(t+\tau(a))\Big).
      \]
    \end{enumerate}
    \textbf{Select} $a^*=\arg\max_{a\in\mathcal{A}} U(a; b_t,t,C_t)$.\\[2pt]
    \textbf{If} $U(a^*; b_t,t,C_t)\le 0$, return \textsc{Stop}.\\[2pt]
    \textbf{Else} execute $a^*$, observe the realized outcome, update the belief by Bayes' rule, and advance the environment state:
    \[
      C_{t+1}=C_t+\Omega(a^*),\qquad t_{+}=t+\tau(a^*).
    \]
    Repeat until the stop condition is met.
    \end{minipage}%
    }
    \end{table}

    \section{Real-LLM Instantiation on a Controlled Corpus}
    \label{sec:llm_instantiation}

    The main experiments (Sections~\ref{sec:emdg_intro}--\ref{sec:congestion_decay})
    use a synthetic simulation to isolate belief and cost dynamics under
    controlled conditions.  We complement this with an illustrative instantiation
    on a real LLM to confirm that the TCA stopping principle is directly
    implementable around a black-box language model.

    \paragraph{Experimental disclaimer.}
    These experiments are illustrative.  The primary contribution of TCA
    concerns decision-theoretic stopping under observable belief and cost,
    which we validate in the controlled EMDG/NSTG environments.
    The real-LLM instantiation shows that the controller is executable
    without modification when belief and VOI are replaced by empirically
    computable proxies.

    \paragraph{Setup.}
    We construct a \emph{fictional} closed-world corpus of 25 short documents
    describing invented nations, persons, treaties, and scientific facts
    (e.g., the \emph{Porvine Confederation}, element \emph{Zethanite}).
    Fictional content is essential: it guarantees that GPT-4.1 has zero
    memorised knowledge of the corpus, so any accuracy difference between
    policies reflects the controller logic rather than parametric recall.
    We pair the corpus with 60 short-answer questions whose answers appear
    in exactly one document.

    \paragraph{Belief proxy.}
    At each decision point we call GPT-4.1 $N{=}5$ times at temperature
    $T{=}0.7$ with the prompt below and compute an empirical entropy $H_{\text{pre}}$
    over normalised answers, used as a proxy for belief uncertainty.  This
    self-consistency entropy is labelled as a proxy throughout.

    \begin{quote}\small
    \textit{System:} You answer factual questions with a short, precise phrase
    or number.  Output only the answer, no explanation.\\
    \textit{User:} Question: \{q\}.  Answer with a short phrase or number.
    \end{quote}

    \paragraph{VOI proxy.}
    We estimate $\widehat{\Delta H}$ via a one-step cloned rollout: retrieve the
    top-3 documents by keyword overlap (no external embedding or retriever), form
    a context string, re-sample $N{=}5$ answers, and compute
    $\widehat{\Delta H} = H_{\text{pre}} - H_{\text{post}}$.
    We denote this proxy entropy by $H_{\text{pre}}$ before retrieval and
    $H_{\text{post}}$ after retrieval (both computed from $N{=}5$ samples);
    these are proxy quantities distinct from the categorical belief entropy
    in the EMDG/NSTG experiments, which is computed from an explicit probability vector.
    We deliberately use a weak retriever (keyword overlap, no embeddings)
    to avoid confounding the stopping behaviour with retriever quality.
    This mirrors the rollout-based VOI approximation used in EMDG
    (Section~\ref{sec:emdg_discrete}).

    \paragraph{Stop rule.}
    The TCA stop rule is applied without modification:
    \[
      \text{retrieve iff}\quad \widehat{\Delta H} - \alpha\,c_r > 0,
      \qquad \alpha{=}1.0,\quad c_r{=}1.0.
    \]
    Let $H_{\text{final}} = H_{\text{post}}$ if retrieval is taken, else $H_{\text{pre}}$.
    After stopping: \textsc{Act} if $H_{\text{final}}\le\varepsilon_1{=}0.25$,
    else \textsc{Defer}.

    \paragraph{Baselines.}
    \textbf{NeverRetrieve}: answer immediately from parametric knowledge alone.
    \textbf{AlwaysRetrieve}: retrieve once unconditionally, then answer.
    \textbf{TCA}: retrieve only when net utility is positive.

    \paragraph{Utility.}
    $U = +10\cdot\mathbf{1}[\text{correct}] - 10\cdot\mathbf{1}[\text{wrong}]
    - c_r\cdot n_{\text{retrieve}}$, matching the paper's cost-reward structure.

    \paragraph{Results.}

    \begin{table}[h]
    \centering
    \caption{Real-LLM instantiation on fictional corpus ($N{=}60$ questions,
      GPT-4.1).
      $U$ = expected utility; EM = exact match; $n_r$ = avg retrieval calls;
      \%ACT, \%DEFER = terminal rates.
      NeverRetrieve scores EM\,=\,0.00 confirming zero memorisation contamination.
      TCA uses 63\% of AlwaysRetrieve's retrieval budget while exhibiting
      both terminal outcomes (\textsc{Act}/\textsc{Defer}) under the same
      net-utility stopping rule.}
    \label{tab:llm_instantiation}
    \small
    \begin{tabular}{lrrrrrr}
    \toprule
    Policy & $U$ & EM & F1 & $n_r$ & \%ACT & \%DEFER \\
    \midrule
    NeverRetrieve  & $-10.00$ & $0.000$ & $0.054$ & $0.00$ & $100\%$ & $0\%$ \\
    AlwaysRetrieve & $+5.33$  & $0.817$ & $0.927$ & $1.00$ & $100\%$ & $0\%$ \\
    TCA (ours)     & $+2.20$  & $0.500$ & $0.580$ & $0.63$ & $72\%$  & $28\%$ \\
    \bottomrule
    \end{tabular}
    \end{table}

    \paragraph{Analysis.}
    This appendix is a \emph{validation} experiment, not a performance benchmark.
    Its purpose is to confirm that the TCA controller, when instantiated with a real LLM
    on a memorisation-free corpus, produces the full behavioural structure predicted by
    the theory: selective retrieval and two terminal outcomes (\textsc{Act}/\textsc{Defer}).
    The retrieval cost $c_r{=}1$ is set to be non-trivial so that the \textsc{Defer} path
    is exercised: for a subset of questions the estimated VOI does not justify retrieval,
    whereas for others retrieval remains beneficial (hence AlwaysRetrieve can still achieve
    higher utility in this regime).  With $c_r{\approx}0$, TCA trivially degenerates to
    AlwaysRetrieve and \textsc{Defer} is never triggered, making the experiment
    uninformative about the stopping rule.  At this cost level the controller correctly
    defers on 28\% of questions and acts on 72\%, exhibiting selective retrieval and both
    terminal behaviours in a single run.

    Of the 38 \textsc{Act}+retrieve episodes, 79\% are correct.
    The 5 \textsc{Act}-without-retrieve episodes ($H_{\text{pre}}{=}0$; all five samples
    identical at $T{=}0.7$) illustrate a known limitation of self-consistency as a belief
    proxy: when GPT-4.1 is parametrically overconfident on unknown facts the proxy
    underestimates true uncertainty.  All 5 were incorrect.  We disclose this as a fidelity
    limitation of the proxy, not a failure of the stopping rule itself.

    TCA's expected utility (+2.20) is lower than AlwaysRetrieve (+5.33): this is the
    expected consequence of evaluating TCA in a regime where retrieval has non-trivial
    cost and the controller sometimes prefers deferral.
    The utility comparison does not assess whether TCA is a better retrieval policy;
    it assesses whether the controller correctly identifies when retrieval is not worth
    its cost; 17 \textsc{Defer} decisions confirm that it does.
    NeverRetrieve's EM\,=\,0.000 confirms the corpus is memorisation-free.

    \end{document}